\documentclass[sigconf]{acmart}
\usepackage[framemethod=tikz]{mdframed}
\usepackage{multirow}

\AtBeginDocument{%
  }

\copyrightyear{2026}
\acmYear{2026}
\setcopyright{cc}
\setcctype{by}
\acmConference[WSDM '26]{Proceedings of the Nineteenth ACM International Conference on Web Search and Data Mining}{February 22--26, 2026}{Boise, ID, USA}
\acmBooktitle{Proceedings of the Nineteenth ACM International Conference on Web Search and Data Mining (WSDM '26), February 22--26, 2026, Boise, ID, USA}
\acmPrice{}
\acmDOI{10.1145/3773966.3785506}
\acmISBN{979-8-4007-2292-9/2026/02}

\title{A Preliminary Agentic Framework for Matrix Deflation}

\author{Paimon Goulart}
\affiliation{
  \institution{University of California, Riverside}
  \city{Riverside}
  \state{CA}
  \country{USA}
}
\email{pgoul002@ucr.edu}

\author{Evangelos E.~Papalexakis}
\affiliation{
  \institution{University of California, Riverside}
  \city{Riverside}
  \state{CA}
  \country{USA}
}
\email{epapalex@cs.ucr.edu}



\begin{document}

\begin{abstract}
Can a small team of agents peel a matrix apart, one rank-1 slice at a time? We propose an agentic approach to matrix deflation in which a solver Large Language Model (LLM) generates rank-1 Singular Value Decomposition (SVD) updates and a Vision Language Model (VLM) accepts or rejects each update and decides when to stop, eliminating fixed norm thresholds. Solver stability is improved through in-context learning (ICL) and types of row/column permutations that expose visually coherent structure. We evaluate on \textsc{Digits} ($8{\times}8$), \textsc{CIFAR-10} ($32{\times}32$ grayscale), and synthetic ($16{\times}16$) matrices with and without Gaussian noise. In the synthetic noisy case, where the true construction rank $k$ is known, numerical deflation provides the noise target and our best agentic configuration differs by only $1.75$ RMSE of the target. For \textsc{Digits} and \textsc{CIFAR-10}, targets are defined by deflating until the Frobenius norm reaches $10\%$ of the original. Across all settings, our agent achieves competitive results, suggesting that fully agentic, threshold-free deflation is a viable alternative to classical numerical algorithms.
\end{abstract}

\maketitle

\section{Introduction and Methods}

Agentic AI enables foundation models (LLMs/VLMs) to collaborate rather than produce one-shot predictions \cite{wu2023autogenenablingnextgenllm}, and ICL can steer LLMs without training \cite{10825481}. We apply these ideas to matrix deflation, which “peels’’ dominant rank-1 structure to isolate signal, compress data, and expose interpretable patterns \cite{Mackey2009Deflation}. Low-rank assumptions are common in scientific analysis, where matrix factorization, SVD, and tensor decomposition extract sequential components across contexts \cite{10825481}. Classical deflation relies on Frobenius-norm thresholds that are scale-dependent and may not reflect structural similarity \cite{Mackey2009Deflation}. We instead propose a threshold-free agentic framework: a solver LLM proposes rank-1 updates using ICL, a VLM evaluator accepts or rejects them and decides when to stop, and lightweight permutations (none, sort, \textsc{GroupNTeach-Block}) expose block structure. We release code at \url{https://github.com/Pie115/Agentic-Deflation}
 and evaluate across Digits, CIFAR-10, and synthetic matrices.

\begin{figure}[H]
\centering
\includegraphics[width=0.75\linewidth]{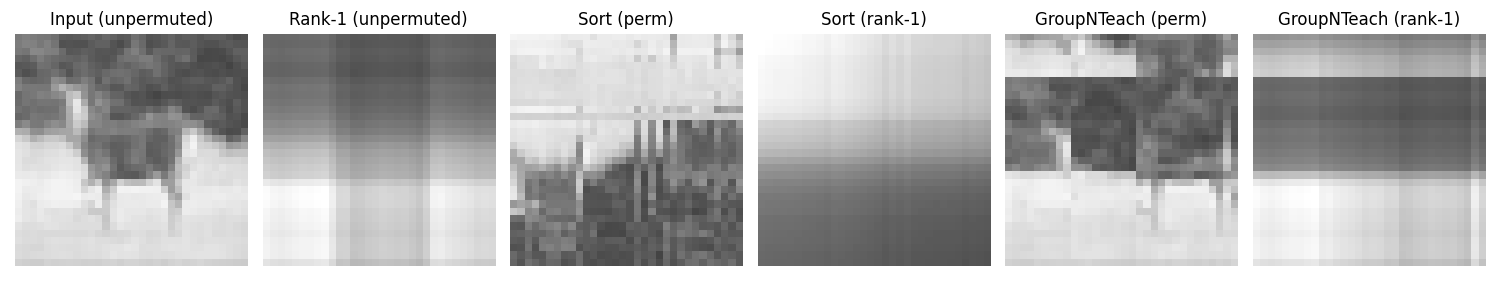}
\caption{\textsc{CIFAR-10} example comparing unpermuted and permuted (Sort, \textsc{GroupNTeach-Block}) rank-1 results.}
\label{fig:cifar_rank1_grid}
\end{figure}

\begin{figure*}[!ht]
\centering
\includegraphics[width=0.98\textwidth]{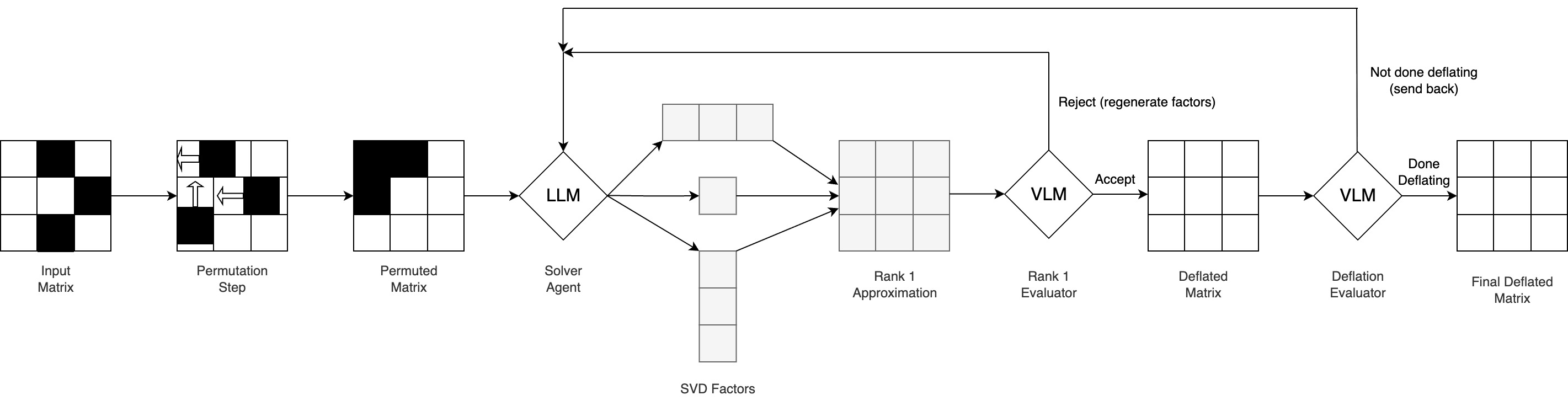}
\caption{LLM-proposed rank-1 updates are VLM-validated and subtracted until deflation stops.}
\label{fig:agentic_framework}
\end{figure*}

\begin{figure}[!ht]
\centering
\includegraphics[width=0.75\linewidth]{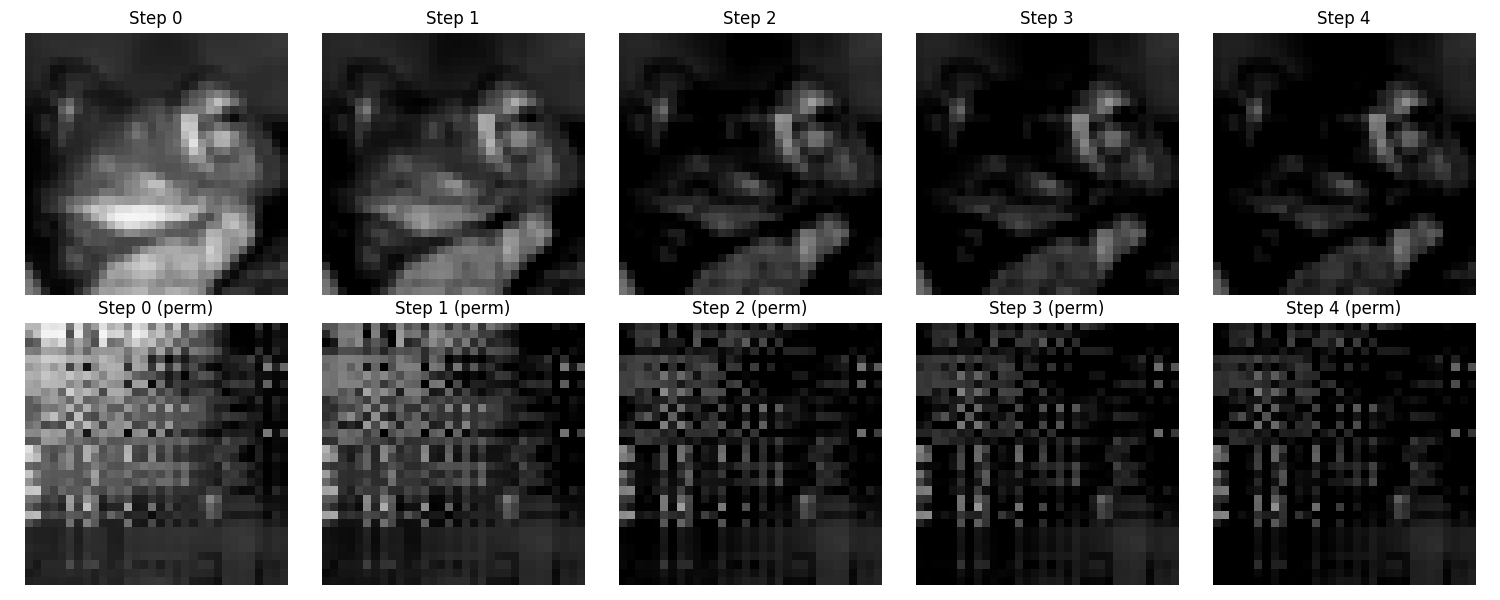}
\caption{Sort (CIFAR-10). Top row = unpermuted view, bottom row = permuted view.}
\label{fig:defl_cifar_sort}
\end{figure}

\textbf{Permutation Preprocessing:} We reorder rows and columns to reveal latent blocky structure \cite{Hooi2016GroupNteach}, making the permuted matrix visually closer to its rank-1 approximation and easier for a VLM to evaluate. In Figure \ref{fig:cifar_rank1_grid} we show different permutations and their corresponding rank-$1$ approximations. \textbf{Solver (Rank-1 proposal via SVD reasoning):} Using Gemini 2.5 Flash \cite{comanici2025gemini25pushingfrontier}, we generate a left singular vector, right singular vector, and scalar value rather than a full $n\times m$ matrix, reducing hallucination and leveraging ICL examples formatted from true SVD decompositions \cite{10825481}. \textbf{SVD Evaluator (Rank-1 quality gate):} Qwen-2-VL \cite{Qwen2VL} compares the permuted input to its reconstructed rank-1 version and either accepts the update or triggers regeneration. \textbf{Deflation Evaluator (Stopping rule):} After updating via $A := \max(0, A - A_1)$, the VLM evaluator inspects the original and current permuted matrices and stops only when meaningful structure is gone, removing the need for a numerical cutoff using the Frobenius norm. \textbf{Agentic Framework:} Figure \ref{fig:agentic_framework} illustrates the full LLM–VLM deflation loop, and Figure \ref{fig:defl_cifar_sort} visualizes the iterative reduction on CIFAR-10.

\section{Experiments}

We evaluate deflation across three datasets: \textsc{Digits} ($8{\times}8$ grayscale), \textsc{CIFAR-10} ($32{\times}32$ natural images) \cite{pen-based_recognition_of_handwritten_digits_81, krizhevsky2009learning}, and a $16{\times}16$ synthetic benchmark where each matrix is generated as $A=\sum_{i=1}^{k} s_i u_i v_i^\top$ with $k\in\{1,\dots,10\}$ and Gaussian noise added before clipping. All inputs are quantized to 8-bit $[0,255]$.

\subsection{Deflation Experiment}

\begin{table}[!ht]
\centering
\scriptsize
\setlength{\tabcolsep}{4pt}
\renewcommand{\arraystretch}{0.9}

\begin{tabular}{lcccc}
\toprule
\textbf{Dataset} & \textbf{Perm} & \textbf{\# ICL} & \textbf{Diff to NumPy} & \textbf{Steps} \\
\midrule
\multirow{3}{*}{Synthetic (Noisy)} 
 & \textsc{GroupNTeach-Block} & 1 & 7.96 & \textbf{10.52} \\
 & None                       & \textbf{2} & \textbf{1.75} & 13.76 \\
 & Sort                       & 4 & 4.69 & 13.54 \\
\midrule
\multirow{3}{*}{Digits} 
 & \textsc{GroupNTeach-Block} & 5 & 21.09 & 3.25 \\
 & None                       & 5 & 26.25 & 3.23 \\
 & \textbf{Sort}              & \textbf{5} & \textbf{16.43} & \textbf{3.75} \\
\midrule
\multirow{2}{*}{CIFAR-10}
 & \textsc{GroupNTeach-Block} & 5 & 50.59 & 1.39 \\
 & \textbf{Sort}              & \textbf{5} & \textbf{31.54} & \textbf{2.39} \\
\bottomrule
\end{tabular}

\caption{Deflation results across all datasets. RMSE difference to the numerical baseline residuals (NumPy) and the number of deflation steps taken by each agent configuration. NumPy baselines: Synthetic (Noisy) — RMSE = 13.46, Steps = 5.6; Digits — Steps = 4.2; CIFAR-10 — Steps = 3.05.}
\label{tab:deflation_all}
\end{table}

On the synthetic (noisy) benchmark, the construction rank $k$ is known, so we deflate numerically for $k$ steps and treat the resulting noise as the target residual. Because $k\sim\mathrm{Unif}\{1,\dots,10\}$, the expected step count is $\mathbb{E}[k]=5.5$. As shown in Table~\ref{tab:deflation_all}, all settings remain close to the numerical baseline, with the smallest RMSE gap using no permutation (ICL=2, $1.75$ above the target). \textsc{GroupNTeach-Block} uses fewer steps (10.52 vs.\ 5.6), reflecting more conservative subtraction.

For \textsc{Digits} and \textsc{CIFAR-10}, true rank is unknown, so we define numerical targets by deflating until the residual Frobenius norm is $10\%$ of the original, then compare RMSE and step counts. On \textsc{Digits}, a simple sort permutation performs best (16.43 RMSE gap, 3.75 vs.\ 4.2 steps). On \textsc{CIFAR-10}, sort again achieves the lowest RMSE gap (31.54) and a step count close to the numerical baseline (2.39 vs.\ 3.05). We omit “None’’ for \textsc{CIFAR-10} because unpermuted matrices repeatedly led to rejected rank-1 proposals, consistent with natural images benefiting from structure-revealing permutations.

\section{Conclusion}
Overall, our agentic framework achieves competitive residuals without norm thresholds, with permutation helping on natural images and conservative updates increasing step counts. In the future, we aim to scale to larger data, tune the agents, incorporate ICL into evaluation, and explore additional permutation strategies.
\section{Ethical Considerations}
We use only synthetic or public image data, posing minimal privacy risk, though applying agentic deflation to user-level behavioral logs would require proper anonymization.
\section{Acknowledgments}
Research was supported in part by a UCR Senate Committee on Research grant, the National Science Foundation under CAREER grant no. IIS 2046086, grant no. No. 2524228  and CREST Center for Multidisciplinary Research Excellence in CyberPhysical Infrastructure Systems (MECIS) grant no. 2112650 and by the Army Research Office and was accomplished under Grant Number W911NF-24-1-0397. The views and conclusions contained in this document are those of the authors and should not be interpreted as representing the official policies, either expressed or implied, of the Army Research Office or the U.S. Government. The U.S. Government is authorized to reproduce and distribute reprints for Government purposes notwithstanding any copyright notation herein.

\bibliographystyle{ACM-Reference-Format}
\bibliography{main}


\begin{thebibliography}{8}


\ifx \showCODEN    \undefined \def \showCODEN     #1{\unskip}     \fi
\ifx \showISBNx    \undefined \def \showISBNx     #1{\unskip}     \fi
\ifx \showISBNxiii \undefined \def \showISBNxiii  #1{\unskip}     \fi
\ifx \showISSN     \undefined \def \showISSN      #1{\unskip}     \fi
\ifx \showLCCN     \undefined \def \showLCCN      #1{\unskip}     \fi
\ifx \shownote     \undefined \def \shownote      #1{#1}          \fi
\ifx \showarticletitle \undefined \def \showarticletitle #1{#1}   \fi
\ifx \showURL      \undefined \def \showURL       {\relax}        \fi
\providecommand\bibfield[2]{#2}
\providecommand\bibinfo[2]{#2}
\providecommand\natexlab[1]{#1}
\providecommand\showeprint[2][]{arXiv:#2}

\bibitem[Alpaydin and Alimoglu(1996)]%
        {pen-based_recognition_of_handwritten_digits_81}
\bibfield{author}{\bibinfo{person}{E. Alpaydin} {and} \bibinfo{person}{Fevzi. Alimoglu}.} \bibinfo{year}{1996}\natexlab{}.
\newblock \bibinfo{title}{{Pen-Based Recognition of Handwritten Digits}}.
\newblock \bibinfo{howpublished}{UCI Machine Learning Repository}.
\newblock
\newblock
\shownote{{DOI}: https://doi.org/10.24432/C5MG6K}.


\bibitem[Comanici et~al\mbox{.}(2025)]%
        {comanici2025gemini25pushingfrontier}
\bibfield{author}{\bibinfo{person}{Gheorghe Comanici} {et~al\mbox{.}}} \bibinfo{year}{2025}\natexlab{}.
\newblock \bibinfo{title}{Gemini 2.5: Pushing the Frontier with Advanced Reasoning, Multimodality, Long Context, and Next Generation Agentic Capabilities}.
\newblock
\showeprint[arxiv]{2507.06261}~[cs.CL]
\urldef\tempurl%
\url{https://arxiv.org/abs/2507.06261}
\showURL{%
\tempurl}


\bibitem[Goulart and Papalexakis(2024)]%
        {10825481}
\bibfield{author}{\bibinfo{person}{Paimon Goulart} {and} \bibinfo{person}{Evangelos~E. Papalexakis}.} \bibinfo{year}{2024}\natexlab{}.
\newblock \showarticletitle{{ Can a Large Language Model Learn Matrix Functions In Context? }}. In \bibinfo{booktitle}{\emph{2024 IEEE International Conference on Big Data (BigData)}}. \bibinfo{publisher}{IEEE Computer Society}, \bibinfo{address}{Los Alamitos, CA, USA}, \bibinfo{pages}{5335--5341}.
\newblock
\href{https://doi.org/10.1109/BigData62323.2024.10825481}{doi:\nolinkurl{10.1109/BigData62323.2024.10825481}}


\bibitem[Hooi et~al\mbox{.}(2016)]%
        {Hooi2016GroupNteach}
\bibfield{author}{\bibinfo{person}{Bryan Hooi}, \bibinfo{person}{Hyun~Ah Song}, \bibinfo{person}{Evangelos~E. Papalexakis}, \bibinfo{person}{Rajiv Agrawal}, {and} \bibinfo{person}{Christos Faloutsos}.} \bibinfo{year}{2016}\natexlab{}.
\newblock \showarticletitle{Matrices, Compression, Learning Curves: Formulation, and the GroupNteach Algorithms}. In \bibinfo{booktitle}{\emph{Advances in Knowledge Discovery and Data Mining (PAKDD)}} \emph{(\bibinfo{series}{Lecture Notes in Computer Science}, Vol.~\bibinfo{volume}{9652})}, \bibfield{editor}{\bibinfo{person}{James Bailey}, \bibinfo{person}{Latifur Khan}, \bibinfo{person}{Takashi Washio}, \bibinfo{person}{Gillian Dobbie}, \bibinfo{person}{Jianzhong Huang}, {and} \bibinfo{person}{Rui Wang}} (Eds.). \bibinfo{publisher}{Springer, Cham}, \bibinfo{pages}{417--429}.
\newblock
\href{https://doi.org/10.1007/978-3-319-31750-2_30}{doi:\nolinkurl{10.1007/978-3-319-31750-2_30}}


\bibitem[Krizhevsky(2009)]%
        {krizhevsky2009learning}
\bibfield{author}{\bibinfo{person}{Alex Krizhevsky}.} \bibinfo{year}{2009}\natexlab{}.
\newblock \bibinfo{booktitle}{\emph{Learning multiple layers of features from tiny images}}.
\newblock \bibinfo{type}{{T}echnical {R}eport}. \bibinfo{institution}{University of Toronto}.
\newblock
\urldef\tempurl%
\url{https://www.cs.toronto.edu/~kriz/learning-features-2009-TR.pdf}
\showURL{%
\tempurl}


\bibitem[Mackey(2009)]%
        {Mackey2009Deflation}
\bibfield{author}{\bibinfo{person}{Lester~W. Mackey}.} \bibinfo{year}{2009}\natexlab{}.
\newblock \showarticletitle{Deflation Methods for Sparse {PCA}}. In \bibinfo{booktitle}{\emph{Advances in Neural Information Processing Systems 21 (NeurIPS 2008)}}, \bibfield{editor}{\bibinfo{person}{Daphne Koller}, \bibinfo{person}{Dale Schuurmans}, \bibinfo{person}{Yoshua Bengio}, {and} \bibinfo{person}{L{\'e}on Bottou}} (Eds.). \bibinfo{publisher}{Curran Associates, Inc.}, \bibinfo{pages}{1017--1024}.
\newblock
\urldef\tempurl%
\url{https://papers.nips.cc/paper/3575-deflation-methods-for-sparse-pca}
\showURL{%
\tempurl}


\bibitem[Wang and et~al.(2024)]%
        {Qwen2VL}
\bibfield{author}{\bibinfo{person}{Peng Wang} {and} \bibinfo{person}{et al.}} \bibinfo{year}{2024}\natexlab{}.
\newblock \showarticletitle{Qwen2-VL: Enhancing Vision-Language Model's Perception of the World at Any Resolution}.
\newblock \bibinfo{journal}{\emph{arXiv preprint arXiv:2409.12191}} (\bibinfo{year}{2024}).
\newblock


\bibitem[Wu et~al\mbox{.}(2023)]%
        {wu2023autogenenablingnextgenllm}
\bibfield{author}{\bibinfo{person}{Qingyun Wu}, \bibinfo{person}{Gagan Bansal}, \bibinfo{person}{Jieyu Zhang}, \bibinfo{person}{Yiran Wu}, \bibinfo{person}{Beibin Li}, \bibinfo{person}{Erkang Zhu}, \bibinfo{person}{Li Jiang}, \bibinfo{person}{Xiaoyun Zhang}, \bibinfo{person}{Shaokun Zhang}, \bibinfo{person}{Jiale Liu}, \bibinfo{person}{Ahmed~Hassan Awadallah}, \bibinfo{person}{Ryen~W White}, \bibinfo{person}{Doug Burger}, {and} \bibinfo{person}{Chi Wang}.} \bibinfo{year}{2023}\natexlab{}.
\newblock \bibinfo{title}{AutoGen: Enabling Next-Gen LLM Applications via Multi-Agent Conversation}.
\newblock
\showeprint[arxiv]{2308.08155}~[cs.AI]
\urldef\tempurl%
\url{https://arxiv.org/abs/2308.08155}
\showURL{%
\tempurl}


\end{thebibliography}
\end{document}